\documentclass[10pt,twocolumn,letterpaper]{article}

\usepackage{cvpr}
\usepackage{times}
\usepackage{epsfig}
\usepackage{graphicx}
\usepackage{amsmath}
\usepackage{amssymb}
\usepackage{multirow}
\usepackage{sidecap}
\usepackage{algorithm}
\usepackage{algorithmic}
\usepackage{flafter}

\usepackage[pagebackref=true,breaklinks=true,letterpaper=true,colorlinks,bookmarks=false]{hyperref}

\cvprfinalcopy 

\def\cvprPaperID{2140}

\ifcvprfinal\fi
\begin{document}

\newcommand{\alg}[1]{Algorithm \ref{alg:#1}}
\newcommand{\eqn}[1]{Eqn.~\ref{eqn:#1}}
\newcommand{\fig}[1]{Fig.~\ref{fig:#1}}
\newcommand{\tab}[1]{Table~\ref{tab:#1}}
\newcommand{\secc}[1]{Section~\ref{sec:#1}}

\def\etal{{\textit{et~al.~}}}

\newcommand{\by}{\ensuremath \times}
\newcommand\patchify{{\tt patchify}}
\newcommand\depatchify{{\tt depatchify}}

\newcommand{\later}[1]{}

\newcommand\enum[1]{({\it #1})}
\newcommand\argmax{\ensuremath \operatornamewithlimits{argmax}}

\newcommand{\deigen}[1]{\textbf{\textcolor{green}{#1}}}
\newcommand{\rob}[1]{\textbf{\textcolor{red}{#1}}}
\newcommand{\li}[1]{\textbf{\textcolor{blue}{#1}}}

\title{End-to-End Integration of a \\ 
    Convolutional Network, Deformable Parts Model \\
    and Non-Maximum Suppression 
\vspace{-5mm}}

\author{Li Wan \quad\quad David Eigen \quad\quad Rob Fergus \\
Dept. of Computer Science, Courant Institute, New York University \\
{\tt\small {wanli,deigen,fergus}@cs.nyu.edu}
}

\maketitle
\begin{abstract}
\vspace{-3mm}
  Deformable Parts Models and Convolutional Networks each have
  achieved notable performance in object detection.  Yet these two
  approaches find their strengths in complementary areas: DPMs are
  well-versed in object composition, modeling fine-grained spatial
  relationships between parts; likewise, ConvNets are adept at
  producing powerful image features, having been discriminatively
  trained directly on the pixels.  In this paper, we propose a new
  model that combines these two approaches, obtaining the advantages
  of each. We train this model using a new structured loss function
  that considers all bounding boxes within an image, rather than
  isolated object instances. This enables the non-maximal suppression
  (NMS) operation, previously treated as a separate post-processing stage,
  to be integrated into the model. This allows for discriminative
  training of our combined Convnet + DPM + NMS model in end-to-end
  fashion. We evaluate our system on PASCAL VOC 2007 and 2011
  datasets, achieving competitive results on both benchmarks.
\vspace{-5mm}

\end{abstract}

\section{Introduction}
\vspace{-2mm}
Object detection has been addressed using a variety of approaches,
including sliding-window Deformable Parts Models
\cite{Felz10,Zhu10,Girshick10}, region proposal with classification
\cite{Ross13,Uij13}, and location regression with deep learning
\cite{Overfeat,Szegedy13}.  Each of these methods have their own
advantages, yet are by no means mutually
exclusive. 
In particular, structured parts models capture the composition of
individual objects from component parts, yet often use rudimentary
features like HoG \cite{dalal-iccv-05} that throw away much of the
discriminative information in the image.  By contrast, deep learning
approaches \cite{Kriz13,Zeiler13,Overfeat}, based on Convolutional Networks \cite{Lecun98}, extract
strong image features, but do not explicitly model object
composition. Instead, they rely on pooling and large fully
connected layers to combine information from spatially disparate
regions; these operations can throw away useful fine-grained spatial
relationships important for detection.

\begin{figure}[ht]
{
\centering
\includegraphics[width=\linewidth]{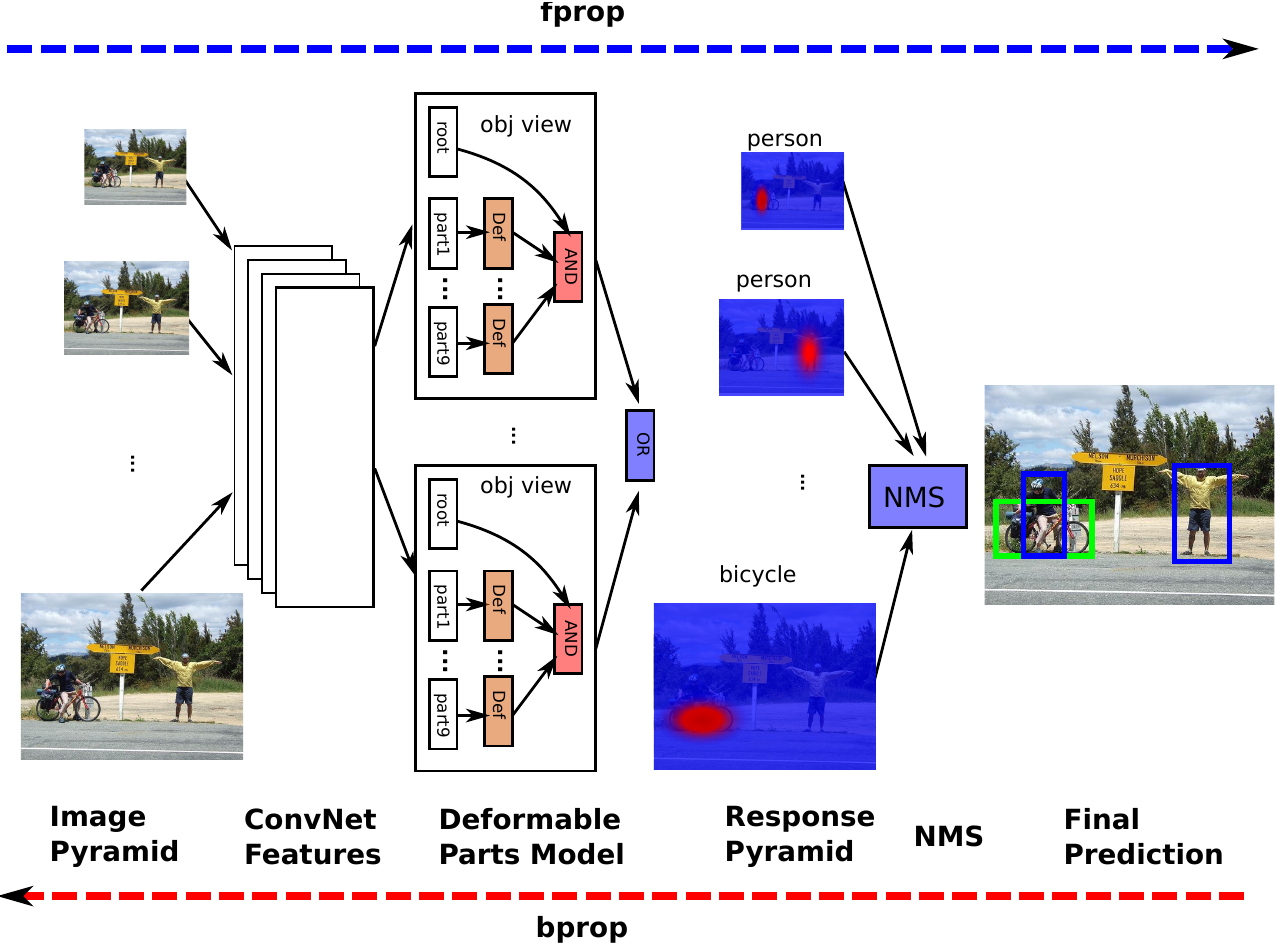}\\
}
\caption{ An overview of our system: (i) a convolutional network 
  extracts features from an image pyramid; (ii) a set of deformable parts
  models (each capturing a different view) are applied to the convolutional
  feature maps; (iii) non-maximal suppression is applied to the
  resulting response maps, yielding bounding box predictions. 
  Training is performed using a new loss function that enables
  back-propagation through all stages.}
\label{fig:overview}
\end{figure}

In this paper, we propose a framework (shown in \fig{overview}) that
combines these two approaches, fusing together structured learning and
deep learning to obtain the advantages of each.  We use a DPM for
detection, but replace the HoG features with features learned by a
convolutional network.  This allows the use of complex image features,
but still preserves the spatial relationships between object parts
during inference. 

\begin{figure}[t]
\centering
\includegraphics[width=0.9\linewidth]{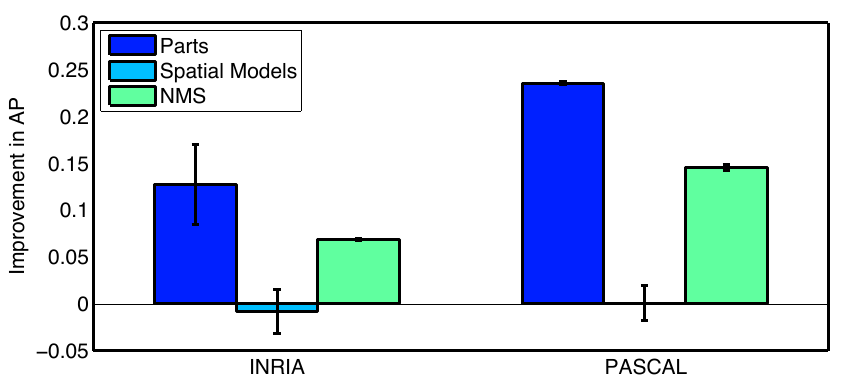}
\caption{Reproduced from Parikh and Zitnick \cite{Parikh11}: an
  ablation study of the stages in a DPM model \cite{Felz10} . Their
  figure shows how significant performance improvements could be obtained by
  replacing the parts detection and non-maximal suppresssion stages
  with human subjects. This suggests that these stages limit
  performance within the model. Our work focuses on improving each of
  these, replacing the part detectors with a Convnet and integrating
  NMS into the model.}
\label{fig:nms-analysis}
\end{figure}

An often overlooked aspect of many detection systems is the
non-maximal suppression stage, used to winnow multiple high scoring
bounding boxes around an object instance down to a single
detection. Typically, this is a post-processing operation applied to
the set of bounding boxes produced by the object detector. As such, it
is not part of the loss function used to train the model and any 
parameters must be tuned by hand. However, as demonstrated by Parikh
and Zitnick \cite{Parikh11}, NMS can be a major performance bottleneck
(see \fig{nms-analysis}). We introduce a new type of image-level loss
function for training that takes into consideration of all bounding boxes 
within an image. This differs with the losses used in existing frameworks that consider
single cropped object instances. Our new loss function enables the
NMS operation trained as part of the model, jointly with the Convnet and DPM components.

\section{Related Work}
Most closely related is the concurrent work of Girshick
\etal~\cite{Girshick14}, who also combine a DPM with ConvNet features in a model
called DeepPyramid DPM (DP-DPM). Their work, however, is
limited to integrating fixed pretrained ConvNet features with a DPM.  We
independently corroborate the conclusion that using ConvNet features in place
of HoG greatly boosts the performance of DPMs.  Furthermore, we show how
using a post-NMS online training loss improves response ordering and addresses
errors from the NMS stage.  We also perform joint end-to-end training of the
entire system.

The basic building blocks of our model architecture come from the DPMs of
Felzenszwalb \etal \cite{Felz10} and Zhu \etal~\cite{Zhu10}\cite{Chen10}, 
and the ConvNet of Krizhevsky \etal \cite{Kriz13}.  We make crucial modifications in their
integration that enables the resulting model to achieve competitive object
detection performance.  In particular, we develop ways to transfer the ConvNet
from classification to the detection environment, as well as changes to the
learning procedure to enable joint training of all parts.

The first system to combine structured learning with a ConvNet is LeCun
\etal\cite{Lecun98}, who train a ConvNet to classify individual digits, 
then train with hand written strings of digits discriminatively.
Very recently, Tompson \etal\cite{Tompson14} trained a
model for human pose estimation that combines body part location estimates into
a convolutional network, in effect integrating an MRF-like model with a ConvNet.
Their system, however, requires annotated body part locations and is applied to
pose estimation, whereas our system does not require annotated parts and is
applied to object detection.

Some recent works have applied ConvNets to object detection directly:  Sermanet
\etal \cite{Overfeat} train a network to regress on object bounding box
coordinates at different strides and scales, then merge predicted boxes across
the image.  Szegedy \etal \cite{Szegedy13} regress to a bitmap with the bounding
box target, which they then apply to strided windows.  Both of these approaches
directly regress to the bounding box from the convolutional network features,
potentially ignoring many important spatial relationships.  By contrast, we use
the ConvNet features as input to a DPM.  In this way, we can include a model of
the spatial relationships between object parts.

In the R-CNN model, Girshick \etal \cite{Ross13} take a different approach in
the use of ConvNets.  Instead of integrating a location regressor into the
network, they instead produce candidate region proposals with a separate
mechanism, then use the ConvNet to classify each region.  However, this
explicitly resizes each region to the classifier field of view (fixed size), performing
significant distortions to the input, and requires the entire network stack to be
recomputed for each region.  Instead, our integration runs the features in a
convolutional bottom-up fashion over the whole image, preserving the true
aspect ratios and requiring only one computational pass.

End-to-end training of a multi-class detector and post-processing has also been
discussed in Desai \etal~\cite{Desai11}.  
Their approach reformulates NMS as a contextual 
relationship between locations.  
They replace NMS, which removes
duplicate detections, with a greedy search that adds detection results
using an object class-pairs context model. 
Whereas their system handles interactions between different types of
objects, our system integrates NMS in a way that 
creates an ordering of results both of different classes and the
same class but different views.  In addition, we further
integrate this into a full end-to-end system including ConvNet feature generation.

\section{Model Architecture}

\begin{figure}[ht]
\centering
\includegraphics[width=0.9\linewidth]{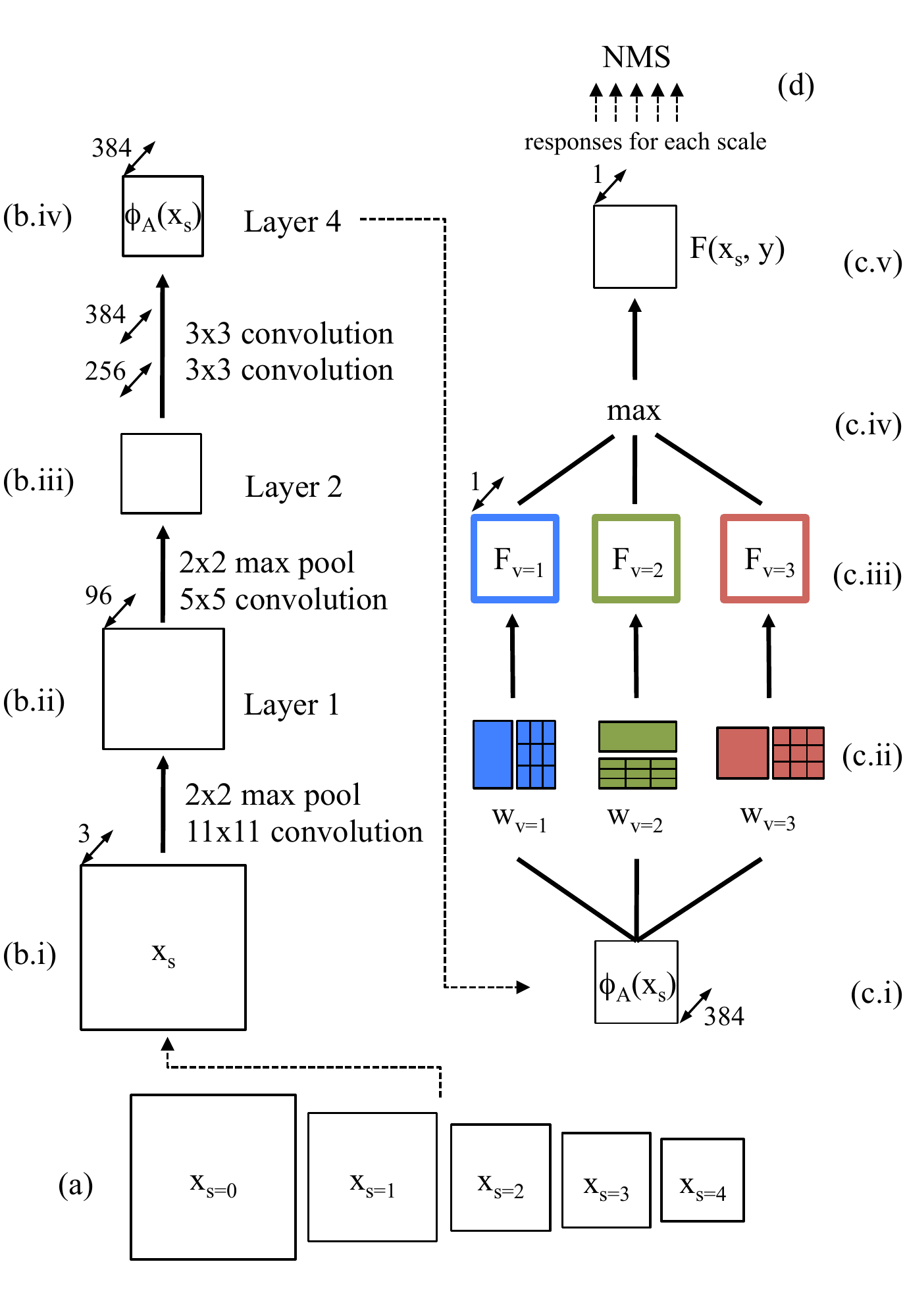}

\caption{Our model architecture, with Convolutional Network (left), 
Deformable Parts Model (right) and non-maximal suppression (top) components.  An input $x$ is first repeatedly
downsampled to create an image pyramid (a).  We run the convolutional network
on each scale, by performing four layers of convolution and max-pooling operations (b.ii - b.iv).  This produces a set of appearance features $\phi_A(x_s)$ at
each scale, which are used as input to a DPM (c.i).  Each object class model
has three views of object templates (c.ii), each of which is composed of
a root filter and nine parts filters.  These produce a response map $F_v$ for each view (c.iii), which are then combined using a pixel-wise max (c.iv)
to form a final activation map for the object class, $F(x_s, y)$.  We then
perform NMS (d) across responses for all scales. To generate
bounding boxes, we trace
the activation locations back to their corresponding boxes in the
input.
}

\label{fig:model}
\end{figure}
The
architecture of our model is shown in \fig{model}.  For a given input image $x$, we first
construct an image pyramid (a) with five intervals over one
octave\footnote{We use as many octaves as required to make the
  smallest dimension 48 pixels in size.} 
We apply the ConvNet (b) at each scale $x_s$ to generate feature maps
$\phi_A(x_s)$.  These are then passed to the DPM (c) for each class; as we
describe in \secc{dpm}, the DPM may also be formulated as a series of neural
network layers.
At training time, the loss is computed using the final detection output obtained after NMS (d),
and this is then back-propagated end-to-end through the entire system, including
NMS, DPM and ConvNet.

\subsection{Convolutional Network}

\label{sec:convnet}

We generate appearance features $\phi_A(x)$ using the first five layers of a
Convolutional Network pre-trained for the ImageNet Classification task. 
We first train an eight layer classification
model, which is composed of five convolutional feature extraction layers, plus
three fully-connected classification layers\footnote{The fully connected layers
have 4096 - 4096 - 1000 output units each, with dropout applied to the two
hidden layers.  We use the basic model from \cite{Zeiler13}, 
which trains the network using random
224x224 crops from the center 256x256 region of each training image, rescaled so the
shortest side has length 256.  This model achieves a top-5 error rate
of 18.1\% on the ILSVRC2012 validation set, voting with 2 flips and 5 translations.}.
After this
network has been trained, we throw away the three fully-connected layers,
replacing them instead with the DPM.  The five convolutional layers are
then used to extract appearance features.

Note that for detection, we apply the convolutional layers to images of arbitrary size (as
opposed to ConvNet training, which uses fixed-size inputs).  Each layer of the
network is applied in a bottom-up fashion over the entire spatial extent of the
image, so that the total computation performed is still proportional to the
image size.  This stands in contrast to \cite{Ross13}, who apply the
ConvNet with a fixed input size to different image regions, and is
more similar to \cite{Overfeat}.

Applying the ImageNet classification model to PASCAL detection has two scale-related problems that
must be addressed.  The first is that there is a total of 16x subsampling
between the input and the fifth layer; that is, each pixel in $\phi_A$ 
corresponds to 16 pixels of input --- this is insufficient for detection, as it
effectively constrains detected bounding boxes to a lie on a 16-pixel grid.
The second
is that the ImageNet classifier was trained on objects that are
fairly large, taking up much of the 224x224 image area.  By
contrast, many target objects in PASCAL are significantly smaller.

To address these, we simply apply the first convolution layer with a stride of
1 instead of 4 when combining with the DPM (however, we also perform 2x2
pooling after the top ConvNet layer due to speed issues in training, making the
net resolution increase only a factor of 2).  This addresses both scale issues
simultaneously.  The feature resolution is automatically increased by
elimination of the stride.  Moreover, the scale of objects presented to the
network at layers 2 and above is increased by a factor of 4, better aligning
the PASCAL objects to the ImageNet expected size 
This is due to the fact that when the second
layer is applied to the output of the stride-1 maps, their field of view is 4x
smaller compared to stride-4, effectively increasing the size of input objects.

Note that changing the stride of the first layer is effectively the same as
upsampling the input image, but preserves resolution in the convolutional
filters (if the filters were downsampled, these would be equivalent operations;
however we found this to work well without changing the filter size, as they
are already just 11x11).

\subsection{Deformable Parts Model}
\label{sec:dpm}

\subsubsection{Part Responses}

\begin{figure}
{
\centering
\includegraphics[width=0.8\linewidth]{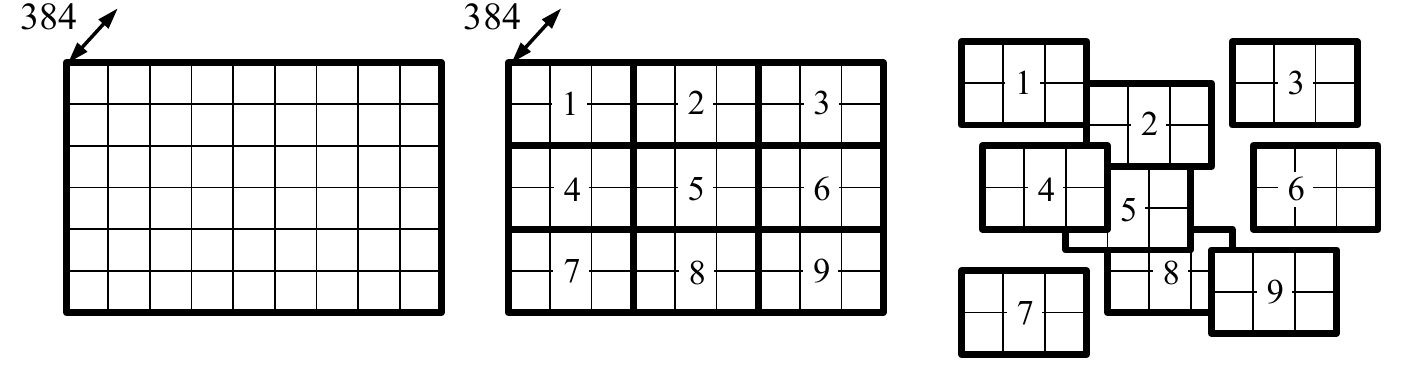}\\
}
\hspace{0.4in}
(a) $w_{A,v}^{\rm root}$
\hspace{0.3in}
(b) $w_{A,v}^{\rm part}$
\hspace{0.3in}
(c) $w_{A,v}^{\rm part}, w_{D,v}^{\rm part}$
\caption{Root and parts filter setup for our DPM.  (a) Each view $v$
  has a root
  filter with a different pre-defined aspect ratio.  (b) Part filters are aligned on a 3x3 grid relative to
the root. (c) Parts may deform relative to the root position at a
cost, parameterized by $w_D^{\text{part}}$.
}
\label{fig:root-parts}
\end{figure}

The first step in the DPM formulation is to convolve the appearance features
with the root and parts filters, producing appearance
responses.
Each object view has both a root filter and nine parts
filters;  the parts are arranged on a 3x3 grid relative to the root, as
illustrated in \fig{root-parts}.  (This is similar to \cite{Zhu10}, who find
this works as well as the more complex placements used by \cite{Felz10}).
Note that the number of root and parts filters is the same for all classes,
but the size of each root and part may vary between classes and views.

Given appearance filters $w_{A,y,v}^{\rm root}$
for each class $y$ and view $v$, and filters $w_{A,y,v,p}^{\rm part}$ for each part $p$,
the appearance scores are:
\begin{eqnarray}
\label{eqn:F}
F_{v}^{\rm root}(x_s, y) & = & w_{A,y,v}^{\rm root} * \phi_A(x_s)
\\
F_{v,p}^{\rm part}(x_s, y) & = & w_{A,y,v,p}^{\rm part} * \phi_A(x_s)
\end{eqnarray}
Part responses are then fed to the deformation layer.

\subsubsection{Deformation Layer}
The deformation layer finds the optimal part locations, accounting for both
apperance and a deformation cost
that models the spatial relation of the part to the root.
Given appearance scores $F_{v,p}^{\rm part}$,
part location $p$ relative to the root, and deformation parameters $w_{D,v,p}$
for each part, the deformed part responses are the following (input variables $(x_s,y)$ omitted):
\begin{equation}
\label{eqn:Fdef}
F_{v,p}^{\rm def}
 =  \max_{\delta_i, \delta_j} ~ 
  F_{v,p}^{\rm part}[p_i + \delta_i, p_j + \delta_j] 
  +  w_{D,v,p}^{\rm part} \phi_D(\delta_i, \delta_j) 
\end{equation}
where $F_{v,p}^{\rm part}[p_i + \delta_i, p_j + \delta_j]$ is the part response map
$F_{v,p}^{\rm part}(x_s, y)$
shifted by spatial offset $(p_i + \delta_i, p_j + \delta_j)$, and $\phi_D(\delta_i, \delta_j)
= [|\delta_i|, |\delta_j|, \delta_i^2, \delta_j^2]^T$ is the
shape deformation feature. $w_{D,y,v}^{\rm part} \ge 0$ are the deformation weights.

Note the maximum in \eqn{Fdef} is taken independently at each output spatial
location: i.e. for each output location, we find the max over possible
deformations $(\delta_i, \delta_j)$.  In practice, searching globally
is unnecessary, and we constrain to search over a window
$[-s,s]\times[-s,s]$ where $s$ is the spatial size of the part (in feature
space).  During training, we save the optimal $(\hat\delta_i,\hat\delta_j)$ at
each output location found during
forward-propagation to use during back-propagation.

The deformation layer extends standard max-pooling over $(\delta_i,\delta_j)$
with \enum{i} a shift offset $(p_i,p_j)$ accounting for the part location, and
\enum{ii} deformation cost $w_D^T \phi_D(\delta_i, \delta_j)$. Setting both of
these to zero would result in standard max-pooling.

\subsubsection{AND/OR Layer}

\begin{figure}
{
\centering
\hspace{2mm}
\includegraphics[width=0.45\linewidth]{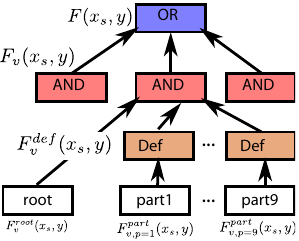}
\hspace{4mm}
\includegraphics[width=0.45\linewidth]{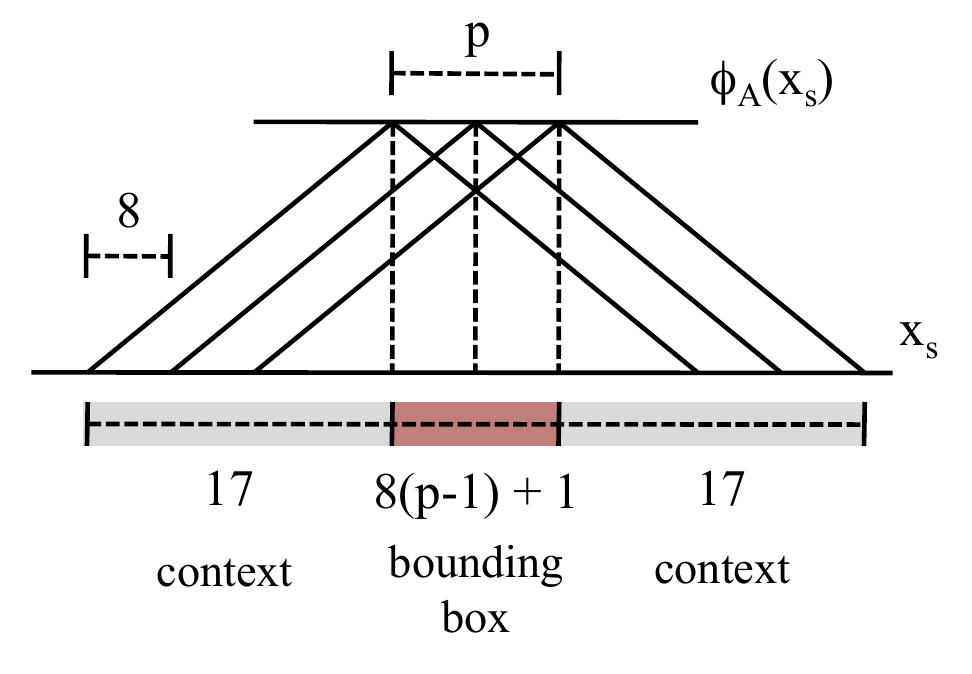}
\vspace{-3mm}
}
\caption{{\it Left}: Overview of the top part of our network architecture: (i) the
  root and part layers are convolution
layers with different sizes of filter but same input size; (ii) OR/AND/Def layer preserve
the size of the input to the output; (iii) each AND layer represents an object 
view which contains a root and 9 parts. \newline
{\it Right}: Aligning a bounding box from DPM prediction through the
convolutional network (see \secc{bbox}). 
}
\label{fig:nn-top}
\end{figure}

Combining the scores of root, parts and object views is
done using an AND-like accumulation over parts to form a score $F_v$ for each view $v$, followed by an OR-like
maximum over views to form the final object score $F$:
\begin{eqnarray}
F_v(x_s, y)\hspace{-1mm} & = &\hspace{-1mm} F^{\rm root}_{v}(x_s, y) + \hspace{-2mm}\sum_{p \in {\rm parts}} \hspace{-2mm}F^{\rm def}_{v,p}(x_s, y) \\
F(x_s, y)  \hspace{-1mm}& = & \hspace{-1mm}\max_{v\in {\rm views}} F_v(x_s, y)
\end{eqnarray}
$F(x_s,y)$ is then the final score map for class $y$ at scale $s$, given the
image $x$ as shown in \fig{nn-top}(left).

\subsection{Bounding Box Prediction}

\label{sec:bbox}

After obtaining activation maps for each class at each scale of input,
we trace the activation locations back to their corresponding bounding
boxes in input space. Detection
locations in $F(x_s,y)$ are first projected back to boxes in
$\phi_A(x_s)$, using the root and parts filter sizes and inferred parts offsets.  These
are then projected back into input space through the convolutional network.  As shown in
\fig{nn-top}(right), each pixel in $\phi_A(x_s)$ has a field of view of 35 pixels in
the input, and moving by 1 pixel in $\phi_A$ moves by 8 pixels in the input (due to the 8x
subsampling of the convolutional model).  Each bounding box is obtained by
choosing the input region that lies between the field of view centers of the box's boundary.
This means that 17 pixels on all sides of the input
field of view are treated as context, and the bounding box is aligned to
the interior region.

\subsection{Non-Maximal Suppression (NMS)}

The procedure above generates a list of label assignments
$A_0=\{(b_i,y_i,r_i)_{i=1\dots |B|}\}$ for the image, where $b_i$
is a bounding box, and $y_i$ and $r_i$ are its associated class label and network response score, \ie
$r_i$ is equal to $F(x, y_i)$ at the output location corresponding to box $b_i$.  $B$ is the set
of all possible bounding boxes in the search.

The final detection result is a subset of this list, obtained by
applying a modified version of non-maximal suppression derived from \cite{Zhu10}.
If we label location $b_i$
as object type $y_i$, some neighbors of $b_i$ might also have received a high scores,
where the neighbors of $b$ are defined as 
$neigh(b)=\{b^\prime | overlap(b,b^\prime)\geq \theta \}$.
However, $neigh(b_i)\setminus b_i$ should not be labeled as $y_i$ to avoid duplicate detections.
Applying this, we get a subset of $A=\{(b_i,y_i,r_i)_{i=1\dots n}\}$ as the final detection 
result; usually $n\ll |B|$. 

When calculating $overlap(b,b')$, we use a symmetric form when the bounding boxes are for
different classes, but an asymmetric form when the boxes are both of the same class.
For different-class boxes, $overlap(b,b^\prime)=\frac{Area( b\cap b^\prime)}{ Area(b\cup b^\prime)}$,
and threshold $\theta=0.75$.  For same-class boxes, \eg boxes of different views or locations,
$overlap(b,b^\prime) =\max\left(\frac{Area(b\cap b^\prime)}{Area(b)},
\frac{Area(b\cap b^\prime)}{Area(b^\prime)}\right)$ and $\theta=0.5$.

\section{Final Prediction Loss}
\label{sec:cost-compare}

\subsection{Motivation}
\begin{SCfigure}
    \centering
    \caption{
        Three possible bounding boxes: 
        blue, green and red (blue closest to the ground
        truth). However, green and red should not be
        considered negative instances (since they may be positive in
        other images where the person is occluded). Thus, we want
         \newline
        $r(blue)>r(red)$ \newline
        $r(blue)>r(green)$
    }
    \includegraphics[width=0.38\linewidth]{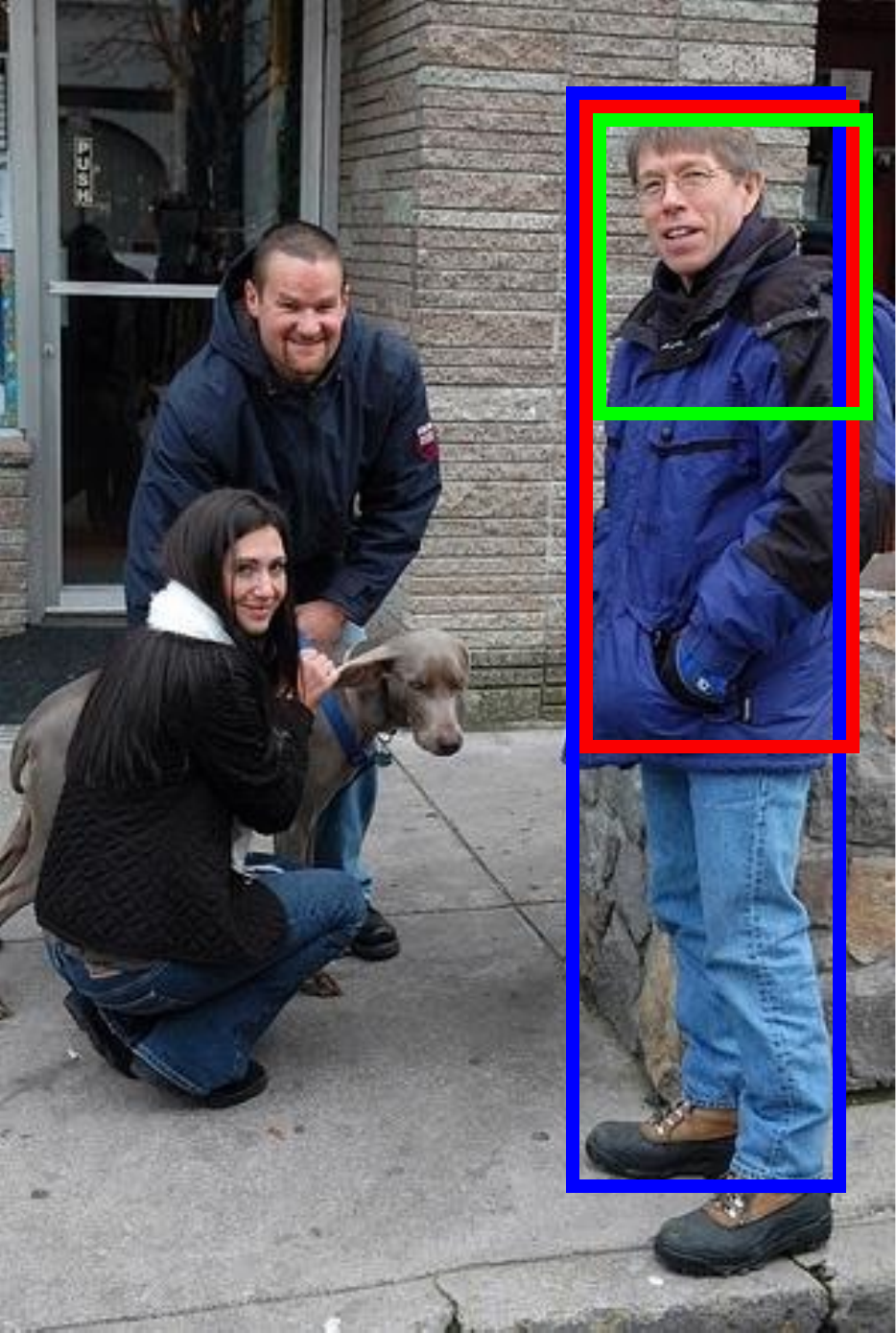}
\label{fig:person}
\end{SCfigure}

Our second main contribution is the use of a final-prediction loss that
takes into account the NMS step used in inference.  In contrast to bootstrapping
with a hard negative pool, such as in \cite{Zhu10}~\cite{Felz10}, we consider each image individually when determining
positive and negative examples, accounting for NMS and the views present in
the image itself.  Consider the example in \fig{person}:
A person detector may fire on three
object views: red, green, and blue. The blue (largest in this example) is closest to the ground 
truth, while green and red are incorrect predictions. However, we cannot simply
add the green or red boxes to a set negative examples, since they are indeed
present in other images as occluded people.  This leads to a
situation where the red view has a higher inference score than blue or green, 
i.e. $r(red)>r(blue)$ and $r(red)>r(green)$, because red is never labeled 
as negative in the bootstrapping process.
After NMS, blue response will be suppressed by red, causing
a NMS error. Such an error can only be avoided when we have
a global view on each image: if $r(blue)>r(red)$, then 
we would have a correct final prediction.

\subsection{Loss Function}
\label{sec:loss_func}

Recall that the NMS stage produces a set of assignments predicted by
the model $A=\{(b_i,y_i,r_i)_{i=1\dots n}\}$ from the set $B$ of all
possible assignments. We compose the loss using two terms, $C(A)$ and $C(A')$.  The first, $C(A)$,
measures the cost incurred by the assignment currently predicted by the model,
while $C(A')$ measures the cost incurred by an assignment close to the ground
truth.  The current prediction cost $C(A)$ is:
\vspace{-4mm}
\begin{equation}
    C(A)=\underbrace{ \sum_{(b_i,y_i,r_i)\in A} \hspace{-3mm} H(r_i,y_i) }_{C^P(A)}+ 
    \underbrace{\sum_{(b_j,y_j,r_j)\in S(A)} \hspace{-2mm} H(r_j,0)}_{C^N(A)}
    \label{eqn:cost-free}
\end{equation}
where 
$H(r,y)=I(y>0)\max(0,1-r)^2 + I(y=0)\max(0,r+1)$
i.e. a squared hinge error.
\footnote{
where $I$ is an indicator function that equals 1 iff the condition holds
}
$S(A)$ is the set of all bounding boxes predicted to be in the background ($y=0$):
$S(A)=B\setminus neigh(A)$ with
$ neigh(A) = \bigcup_{(b_i,y_i,r_i)\in A} neigh(b_i)$. $C_P(A)$ and
$C_N(A)$ are the set of positive predicted labels and the set of
background labels, respectively. 

The second term in the loss, $C(A^\prime)$, measures the cost incurred by the 
ground truth bounding boxes under the model.
Let the ground truth bounding box set be $A^{gt}=\{(b^{gt}_i,y^{gt}_i)_{i=1\dots m}\}$. 
We construct a constrained inference assignment 
$A'$ close to $A^{gt}$
by choosing for each $b^{gt}_i$ the box
$(b'_i,y'_i,r'_i) = \argmax_{(b,y,r)} r$, where the $\argmax$ is taken over all $overlap(b,b^{gt}_i)\geq \theta'$; that is,
the box with highest response out of those with sufficient overlap with the ground truth.
($\theta^\prime=0.7$ in our experiments.)
Similarly to before, the cost $C(A^\prime)$ is:
\begin{equation}
    \label{eqn:cost-constrain}
    C(A^\prime)=\sum_{(b'_i,y'_i,r'_i)\in A'} \hspace{-4mm} H(r'_i,y'_i) + 
    \hspace{-3mm}
    \sum_{(b'_j,y'_j,r'_j)\in S(A')} \hspace{-5mm} H(r'_j,0)
\end{equation}

Thus we measure two costs: that of the current model prediction, and that of
an assignment close to the ground truth.
The final discriminative training loss is difference between these two:
\begin{equation}
L(A,A') = C(A^\prime)-C(A)\geq 0
\end{equation}
Note this loss
is always greater than 0 because the constrained assignment always has cost
at least as large as the unconstrained one, and
$L(A,A') = 0$ when $A = A'$, i.e. when we produce
detection results which are consistent with the ground truth $A^{gt}$.

\newcommand{\bbtup}[1]{\ensuremath {(b#1,y#1,r#1)}}
\newcommand{\bbtupprime}[1]{\ensuremath {(b'#1,y'#1,r'#1)}}

\begin{figure}[!th]
{
\centering
\includegraphics[width=0.8\linewidth]{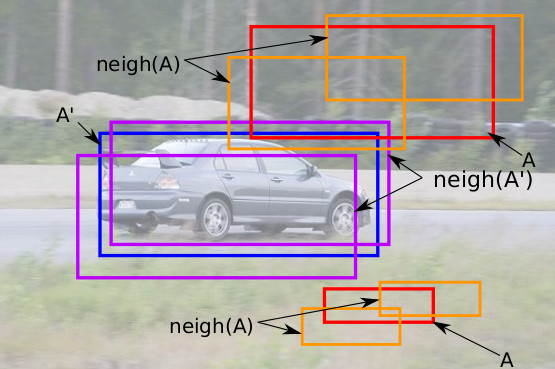}\\
}
\caption{Illustration of ground-truth-constrained assignment $A'$ and
         unconstrained assignments $A$ from the model,
         along with associated neighborhoods.  Note neighborhoods are actually
         dense, and we show only a few boxes for illustration.
         }
\label{fig:nms-loss}
\end{figure}

Combining Equations~\ref{eqn:cost-free} and~\ref{eqn:cost-constrain} leads to
\begin{eqnarray}
    L(A,A') &=& L^P(A,A') + L^N (A,A') \label{eqn:cost-all} \\
    L^P(A,A') &=& \hspace{-5mm} \sum_{\bbtupprime{}\in A'} \hspace{-2mm} H(r',y')
                - \hspace{-3mm} \sum_{\bbtup{}\in A} \hspace{-2mm} H(r, y)
                \nonumber \\
    L^N(A,A')&=& \hspace{-5mm} \sum_{\bbtupprime{}\in S(A^\prime)} \hspace{-2mm} H(r',0)
               - \hspace{-3mm} \sum_{\bbtup{} \in S(A)} \hspace{-2mm} H(r, 0)
               \nonumber \\
             &=& \hspace{-5mm} \sum_{\bbtup{}\in N\setminus N'} \hspace{-2mm} H(r,0)
               - \hspace{-3mm}  \sum_{\bbtupprime{}\in N'\setminus N} \hspace{-2mm} H(r',0) \nonumber
\end{eqnarray}
where $N = neigh(A)$ and $N' = neigh(A')$.
The last line comes from the fact that most of the boxes included in
$S(A)$ and $S(A')$ are shared, and cancel out (see \fig{nms-loss});
thus we can compute the loss looking only
at these neighborhood sets.

\subsection{Interpretation and Effect on NMS Ordering}

As mentioned earlier, a key benefit to training on the final predictions as
we describe is that our loss accounts for the NMS inference step.  In our
example in \fig{person}, if the response $r(red) > r(blue)$, then
$red \in A$ and $blue \in A'$.  Thus $L^P(A,A')$ will decrease $r(red)$ and
increase $r(blue)$.  This ensures the responses $r$ are in an appropriate order
when NMS is applied.  Once $r(blue) > r(red)$, the mistake will be fixed.

The term $L^N$ in the loss is akin to an online version of hard negative mining,
ensuring that the background is not detected as a positive example.

\subsection{Soft Positive Assignments}

When training jointly with the ConvNet, it is insufficient to measure the cost using only single
positive instances, as the network can easily overfit to the
individual examples.  We address this using soft positive assignments in place
of hard assignments; that is, we replace the definition of $C^P$ in \eqn{cost-free} 
used above with
one using a weighted average of neighbors for each box in the assignment list:
$$
C^P(A)=\sum_{b_i\in A} \frac{\sum_{b_j\in neigh(b_i)} \alpha_{ij} H( r_j, y_j)}{\sum_j \alpha_{ij}}
$$
where $\alpha_{ij}=2(Area(b_i\cap b_j)/Area(b_i))-1$, and similarly for $C^P(A')$.

Note a similar strategy has been tried in the case of HoG features before, but was
not found to be beneficial \cite{Felz10}.  By contrast, we found this to be
important for integrating the ConvNet.  We believe this is because the ConvNet
has many more parameters than can easily overfit, whereas HoG is more constrained.

\section{Training}

Our model is trained in online fashion with SGD, with each image being forward
propagated through the model and then the resulting error
backpropagated to update the parameters. During the fprop, the
position of the parts in the DPM are computed and then used for the
subsequent bprop. Training in standard DPM models~\cite{Felz10}
differs in two respects: (i) a fixed negative set is mined
periodically (we have no such set, instead processing each image in
turn) and (ii)  part positions on this negative set are fixed for many subsequent
parameter updates.

We first pretrain the DPM root and parts filters without any
deformation, using a fixed set of 20K random negative examples for each class.
Note that during this stage, the ConvNet weights are fixed to their initialization
from ImageNet.
Following this, we perform end-to-end joint training of the entire system,
including ConvNet, DPM and NMS (via the final prediction loss).  During joint
training, we use inferred part locations in the deformation layer.

The joint training phase is outlined in Algorithm 1.  For each training image sample,
we build an image pyramid, and fprop each scale of the pyramid through the 
ConvNet and DPM to generate the assignment list $A_0$.  Note $A_0$ is
represented using
the output response maps.  We then apply NMS to get the final assignments $A$,
as well as construct the ground-truth constrained assignments $A'$.  Using the final prediction
loss $L(A,A')$ from \eqn{cost-all}, we find the gradient and backpropagate through the network
to update the model weights.  We repeat this for 15 epochs through the training set with a
learning rate $\eta=10^{-3}$, then another 15 using $\eta=10^{-4}$.

At test time, we simply forward-propagate the input pyramid through the network
(ConvNet and DPM) and apply NMS.

\begin{algorithm}[ht]
  \label{alg:trainingAlg}
  \caption{Training algorithm for each image}
  \begin{algorithmic}[1]
    \STATE {\bfseries Input: } Image $X$ with ground truth assignment $A^{gt}$
    \STATE Build image pyramid $X\rightarrow X_1,X_2,\dots,X_s$
    \STATE $A_0=\{\}$
    \FOR{ $X_i\in X_1,X_2,\dots X_s$}
       
       \STATE $A_0=A_0\cup \mbox{assignments from responses~}F(X_i;w)$ 
    \ENDFOR
    \STATE find $A = NMS(A_0)$
    \STATE find $A'$ using $A_0$ and $A^{gt}$
    \FOR{ $X_i\in X_1,X_2,\dots X_s$}
       \STATE find gradient at scale $i$: $g_i = \frac{\partial L(A,A')}{\partial F(X_i;w)}\frac{\partial F(X_i;w)}{\partial w}$
    \ENDFOR
    \STATE $w\leftarrow w +\eta \sum_i g_i$
  \end{algorithmic}
\end{algorithm}

\section{Experiments}

We apply our model to the PASCAL VOC 2007 and VOC 2011/2012 object detection tasks
\cite{pascal12}.
Table~\ref{tab:relative-results} shows how each component in our system
improves performance on the PASCAL 2007 dataset.  Our baseline implementation
of HoG DPM with bootstrap training achieves 30.7 mAP.  Switching HoG for a
fixed pretrained ConvNet results in a large 32\% relative performance gain to
40.8 mAP, corroborating the finding of \cite{Girshick14} that such features
greatly improve performance.  On top of this,
training using our online post-NMS procedure improves substantially improves
performance to 43.3 mAP, and jointly training all components (ConvNet + DPM + NMS)
further improves to 46.5 mAP.
\begin{table}[t]
\small{
    \begin{center}
    \begin{tabular}{|c|c|c|c|c|}
    \hline
                 & Bootstrap & NMS loss & NMS loss+FT \\
    \hline
    HoG-root & 22.8 & 23.9 & N/A \\
    \hline
    HoG-root+part & 30.7 & 33.2 & N/A \\
    \hline
    conv-root &  38.7 & 40.3 & 43.1 \\
    \hline
    conv-root+part &  40.8 & 43.3 & 46.5 \\
    \hline
    \end{tabular}
    \end{center}
}
    \caption{
    \small{
      A performance breakdown of our approach. Columns show
      different training methods and loss functions. Rows show different feature
      extractors and DPM with/without parts. Note: (i) conv features
      give a significant boost; (ii) our new NMS loss consistently
      improves performance, irrespective of features/model used and
      (iii) fine-tuning (FT) of the entire model gives further gains.}
    }
    \label{tab:relative-results}
\vspace{-2mm}
\end{table}
In addition, we can train different models to produce detections for each
class, or train all classes at once using a single model with shared ConvNet
feature extractor (but different DPM components).  Training all classes
together further boosts performance to 46.9\% mAP.  Note that this allows the
post-NMS loss to account for objects of different classes as well as locations
and views within classes, and also makes inference faster due to the shared
features.  We call this model ``conv-dpm+FT-all'', and the separate-class
set of models ``conv-dpm+FT''.

Comparisons with other systems are shown in 
Tables \ref{tab:pascal07-results} (VOC 2007) and \ref{tab:pascal11-results} (VOC 2011/2012).
For VOC 2007 (\tab{pascal07-results}),
our results are very competitive, beating all other methods except the latest version of R-CNN trained
on $fc_7$ (``R-CNN(v4)FT $fc_7$'').  
Notably, we outperform the
DP-DPM method ($45.2\%$ vs. our $46.9\%$), due to our integrated joint training and online NMS loss.
In addition, our final model achieves comparible performance to R-CNN~\cite{Ross13}
with a similar feature extractor using $pool_5$ features ($46.9\%$ vs. $47.3\%$).
Recent version of R-CNN achieve 
a better performance $54.2\%$ using a more complex network which includes
fully connected layers ($fc_7$); extending our model to use deeper networks may also
provide similar gains from better feature representations.

\tab{pascal11-results} shows our system 
performance on VOC2011.  Here our system outperforms comparison methods, 
and in particular DP-DPM, achieving $43.7\%$ mAP versus
$29.6\%$ for HoG-DPM~\cite{Felz10} and $41.6\%$ for DP-DPM~\cite{Girshick14}.

\begin{figure}[ht]
\centering
\includegraphics[width=0.8\linewidth]{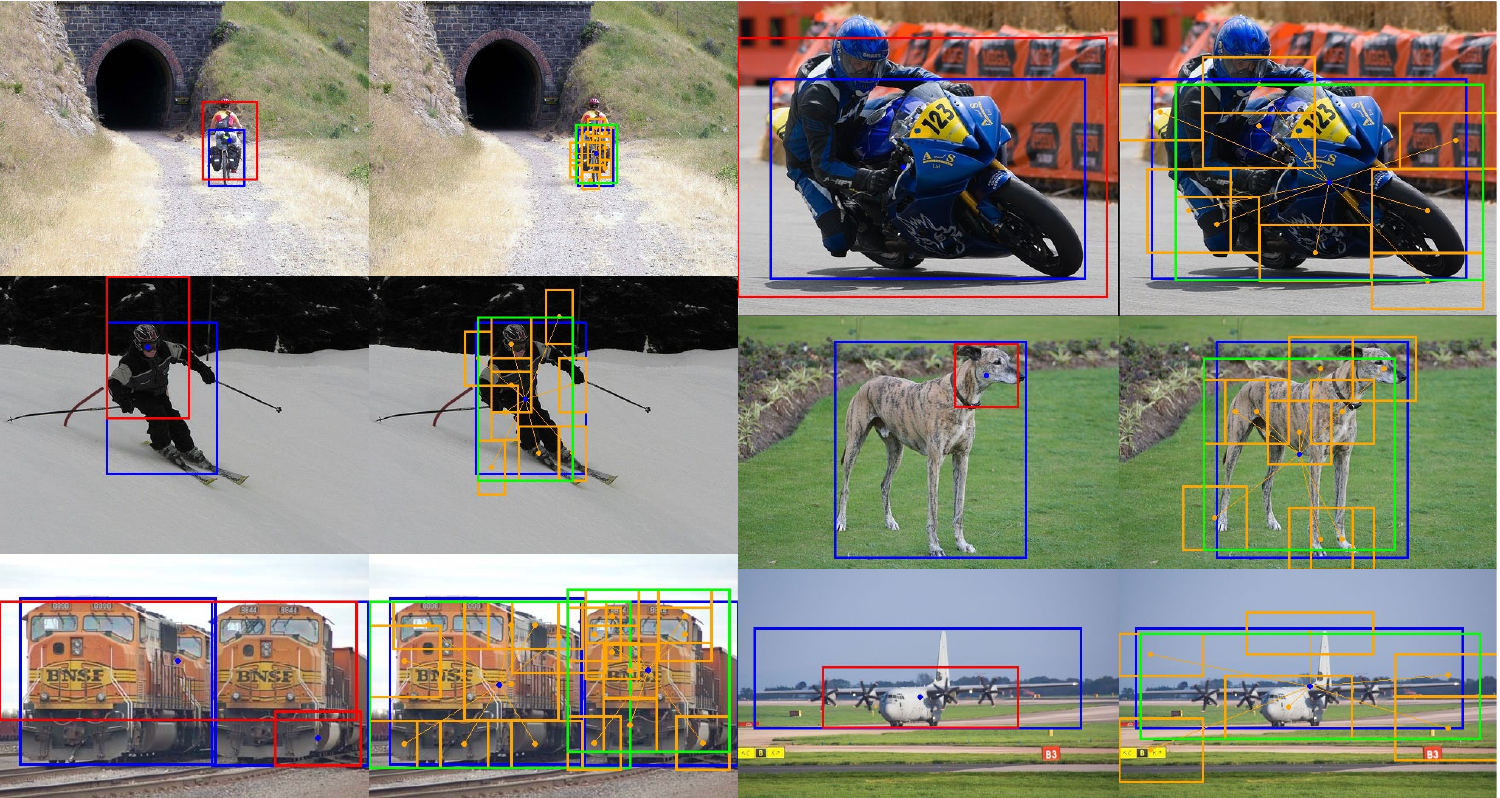}
\caption{
Examples of detections using root filter only (left half of each example; red)
and the DPM with both root and part filters (right halves; green+orange).  
}
\label{fig:examples-convdpm}
\end{figure}

\begin{figure*}[!htbp]
\centering
\includegraphics[width=0.85\linewidth]{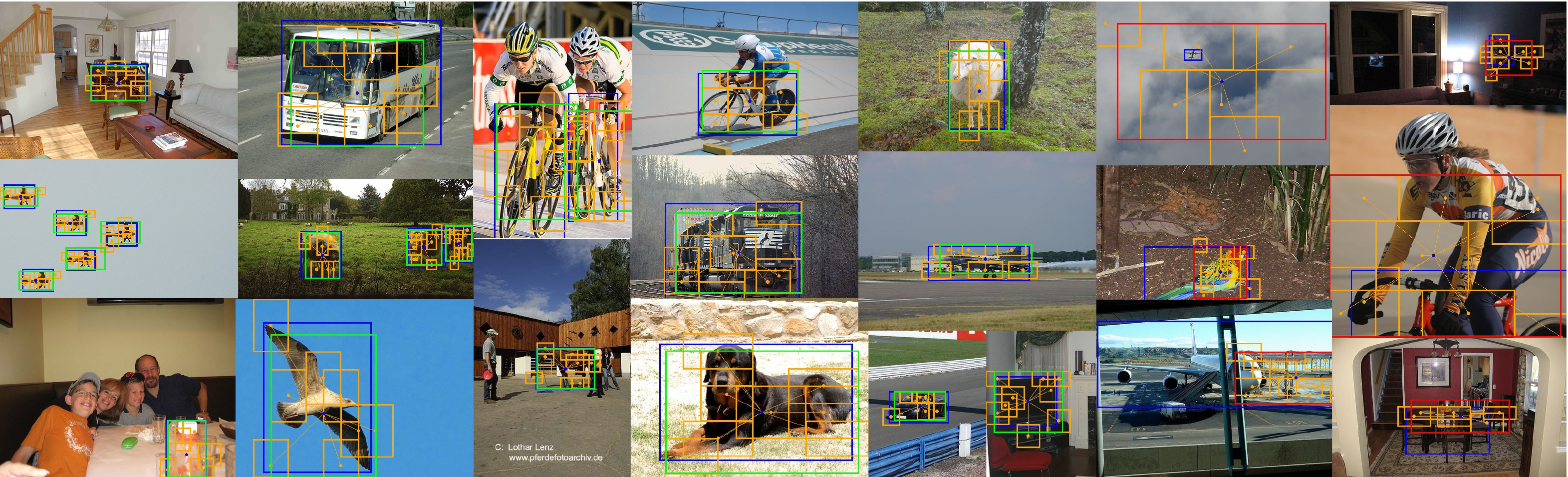}
\caption{
Examples of correct (green) and incorrect (red) detections found by our model.
}
\label{fig:examples-det}
\end{figure*}

\begin{figure*}[!htbp]
\centering
\includegraphics[width=0.85\linewidth]{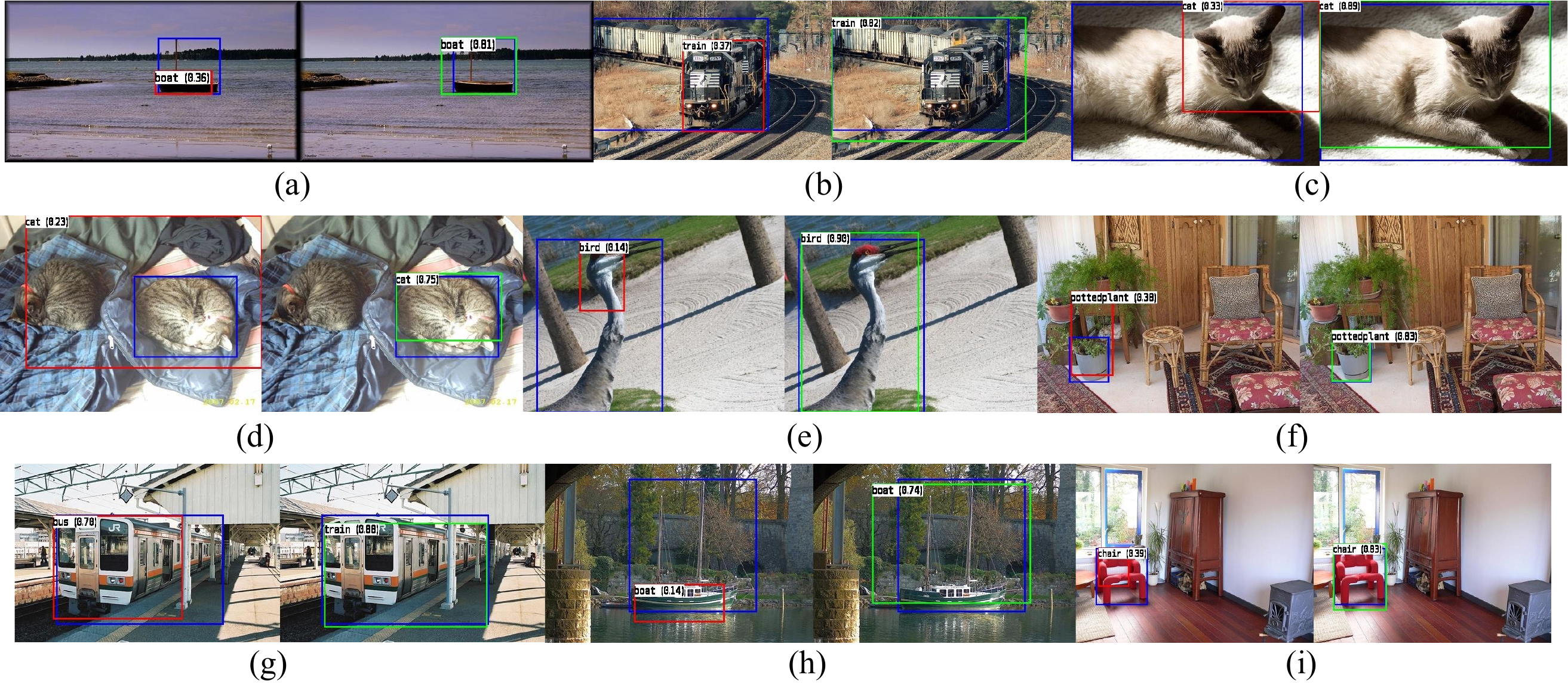}
\caption{
Examples of model with (green) and without (red) NMS loss (parts location are ommited)
}
\label{fig:examples-nms}
\end{figure*}

Finally, we provide examples of detections from our model in
Figures~\ref{fig:examples-convdpm},~\ref{fig:examples-det}
and~\ref{fig:examples-nms}.  Detection results are either show in green or red
with ground truth bounding box in blue.  Figure~\ref{fig:examples-nms}
illustrates training with our new loss function helps model fix problem for
both inter-class and intra-class NMS.  Our loss allows the larger view of the
train to be selected in $(b)$, rather than the more
limited view that appears in more images.  However, the gains are not limited
to selecting larger views:  In $(d)$, we see a cat
correctly selected at a smaller scale.  Finally, there are also examples of
inter-class correction in $(g)$, \eg ``train'' being selected over ``bus''.

Figure~\ref{fig:examples-convdpm} shows the effect of using a DPM with 
parts over just the root-only model. 
Figures~\ref{fig:examples-det} shows correct and incorrect detection examples.

\vspace{-5mm}
\section{Discussion}

We have described an object detection system that integrates a Convolutional Network,
Deformable Parts model and NMS loss in an end-to-end fashion.  This fuses together aspects from both
structured learning and deep learning: object structures are modeled by a
composition of parts and views, while discriminative features are leveraged for
appearance comparisons.  Our evaluations show that our model achieves
competitive performance on PASCAL VOC 2007 and 2011 datasets, and
achieves substantial gains from integrating both ConvNet features as well as NMS, and training all parts jointly.

\begin{table*}[ht]
\begin{center}
    \scalebox{.7}{ 
    \tabcolsep=0.11cm
\begin{tabular}{|l|cccccccccccccccccccc|c|}
        \hline
        VOC2007 & aero & bike & bird & boat & botl & bus & car & cat & chair & cow & table & dog & horse & mbike & pers & plant & sheep & sofa & train & tv & mAP \\
        \hline
        DetectorNet~\cite{Szegedy13} & 29.2 & 35.2 & 19.4 & 16.7 & 3.7 & 53.2 & 50.2 & 27.2 & 10.2 & 34.8 & 30.2 & 28.2 & 46.6 & 41.7 & 26.2 & 10.3 & 32.8 & 26.8 & 39.8 & 47.0 & 30.5 \\
        HoG-dpm(v5)~\cite{Felz10} & 33.2 & 60.3 & 10.2 & 16.1 & 27.3 & 54.3 & 58.2 & 23.0 & 20.0 & 24.1 & 26.7 & 12.7 & 58.1 & 48.2 & 43.2 & 12.0 & 21.1 & 36.1 & 46.0 & 43.5 & 33.7 \\
        HSC-dpm~\cite{Ren13} & 32.2 & 58.3 & 11.5 & 16.3 & 30.6 & 49.9 & 54.8 & 23.5 & 21.5 & 27.7 & 34.0 & 13.7 & 58.1 & 51.6 & 39.9 & 12.4 & 23.5 & 34.4 & 47.4 & 45.2 & 34.3 \\
        Regionlets~\cite{Wang13} & 54.2 & 52.0 & 20.3 & 24.0 & 20.1 & 55.5 & 68.7 & 42.6 & 19.2 & 44.2 & 49.1 & 26.6 & 57.0 & 54.5 & 43.4 & 16.4 & 36.6 & 37.7 & 59.4 & 52.3 & 41.7 \\
        DP-DPM~\cite{Girshick14} & 44.6 & 65.3 & 32.7 & 24.7 & 35.1 & 54.3 & 56.5 & 40.4 & 26.3 & 49.4 & 43.2 & 41.0 & 61.0 & 55.7 & {\bf 53.7} & 25.5 & 47.0 & 39.8 & 47.9 & {\bf 59.2} & 45.2 \\
        \hline
        R-CNN~\cite{Ross13}$fc_7$ & 56.1 & 58.8 & 34.4 & 29.6 & 22.6 & 50.4 & 58.0 & 52.5 & 18.3 & 40.1 & 41.3 & {\bf 46.8} & 49.5 & 53.5 & 39.7 & 23.0 & 46.4 & 36.4 & 50.8 & 59.0 & 43.4 \\
        R-CNN(v1)FT $pool_5$ & 55.6 & 57.5 & 31.5 & 23.1 & 23.2 & 46.3 & 59.0 & 49.2 & 16.5 & 43.1 & 37.8 & 39.7 & 51.5 & 55.4 & 40.4 & 23.9 & 46.3 & 37.9 & 49.7 & 54.1 & 42.1 \\
        R-CNN(v4)FT $pool_5$ & 58.2 & 63.3 & {\bf 37.9} & 27.6 & 26.1 & 54.1 & {\bf 66.9} & {\bf 51.4} & {\bf 26.7} & {\bf 55.5} & 43.4 & 43.1 & 57.7 & 59.0 & 45.8 & {\bf 28.1} & {\bf 50.8} & 40.6 & 53.1 & 56.4 & {\bf 47.3} \\
        R-CNN(v1)FT $fc_7$  & 60.3 & 62.5 & 41.4 & 37.9 & 29.0 & 52.6 & 61.6 & 56.3 & 24.9 & 52.3 & 41.9 & 48.1 & 54.3 & 57.0 & 45.0 & 26.9 & 51.8 & 38.1 & 56.6 & 62.2 & 48.0 \\
        R-CNN(v4)FT $fc_7$ & {\bf 64.2} & {\bf 69.7} & {\bf 50.0} & {\bf 41.9} & 32.0 & {\bf 62.6} & {\bf 71.0} & {\bf 60.7} & {\bf 32.7} & {\bf 58.5} & 46.5 & {\bf 56.1} & 60.6 & {\bf 66.8} & {\bf 54.2} & {\bf 31.5} & {\bf 52.8} & {\bf 48.9} & {\bf 57.9} & {\bf 64.7} & {\bf 54.2}\\
        \hline
        HoG & 29.3 & 55.5 & 9.3 & 13.3 & 25.2 & 43.1 & 53 & 20.4 & 18.5 & 25.1 & 23.3 & 10.3 & 55.4 & 44.2 & 40.8 & 10.5 & 19.8 & 34.3 & 43.3 & 39.5 & 30.7 \\
        HoG+ & 32.8 & 58.5 & 10.3 & 16.0 & 27.1 & 46.1 & 56.9 & 21.9 & 20.6 & 27.2 & 26.4 & 13.0 & 57.8 & 47.5 & 44.2 & 11.0 & 22.7 & 36.5 & 45.8 & 42.1 & 33.2 \\
        conv-root & 38.1 & 60.9 & 21.9 & 17.8 & 29.3 & 51.4 & 58.5 & 26.7 & 16.5 & 31.1 & 33.2 & 24.2 & 65.0 & 58.0 & 44.4 & 21.7 & 35.4 & 36.8 & 49.5 & 54.1 & 38.7 \\
        conv-dpm & 45.3 & 64.5 & 21.1 & 21.0 & 34.2 & 54.4 & 59.0 & 32.6 & 20.0 & 31.0 & 34.5 & 25.3 & 63.8 & 60.1 & 45.0 & 23.2 & 36.0 & 38.4 & 51.5 & 56.2 & 40.8 \\
        conv-dpm+ & 48.9 & 67.3 & 25.3 & 25.1 & 35.7 & 58.3 & 60.1 & 35.3 & 22.7 & 36.4 & 37.1 & 26.9 & 64.9 & 62.0 & 47.0 & 24.1 & 37.5 & 40.2 & 54.1 & 57.0 & 43.3 \\
        conv-dpm+ FT & {\bf 50.9} & 68.3 & 31.9 & 28.2 & 38.1 & 61.0 & 61.3 & 39.8 & 25.4 & 46.5 & 47.3 & 29.6 & 67.5 & {\bf 63.4} & 46.1 & 25.2 & 39.1 & 45.4 & 57.0 & 57.9 & 46.5 \\
        conv-dpm+ FT-all & 49.3 &  {\bf 69.5} & 31.9 & {\bf 28.7} & {\bf 40.4} & {\bf 61.5} & 61.5 & 41.5 & 25.5 & 44.5 & {\bf 47.8} & 32.0 & {\bf 67.5} & 61.8 & 46.7 & 25.9 & 40.5 & {\bf 46.0} & {\bf 57.1} & 58.2 & {\bf 46.9}  \\ 
        \hline
\end{tabular}
} 
\end{center}
\caption{Mean AP on PASCAL VOC 2007}
\label{tab:pascal07-results}
\end{table*}

\begin{table*}[ht]
\begin{center}
    \scalebox{.7}{ 
    \tabcolsep=0.11cm
\begin{tabular}{|l|cccccccccccccccccccc|c|}
        \hline
        
        VOC2011/2012 & aero & bike & bird & boat & botl & bus & car & cat & chair & cow & table & dog & horse & mbike & pers & plant & sheep & sofa & train & tv & mAP \\
        \hline
        HoG-DPM~\cite{Felz10} & 45.6 & 49.0 & 11.0 & 11.6 & 27.2 & 50.5 & 43.1 & 23.6 & 17.2 & 23.2 & 10.7 & 20.5 & 42.5 & 44.5 & 41.3 & 8.7 & 29.0 & 18.7 & 40.0 & 34.5 & 29.6 \\

        conv-root & 56.0 & 45.4 & 20.6 & 12.7 & 29.5 & 49.2 & 38.6 & 38.1 & 16.4 & 28.2 & 22.9 & 28.8 & 48.3 & 52.1 & 47.7 & 17.0 & 39.1 & 29.6 & 41.2 & 48.6 & 35.5 \\
        conv-dpm & 56.9 & 53.2 & 26.6 & 17.6 & 29.9 & 51.4 & 42.5 & 42.4 & 16.5 & 31.6 & 25.0 & 37.7 & 52.7 & 56.7 & 49.9 & 16.5 & 41.0 & 30.9 & 44.4 & 49.7 & 38.4 \\
        conv-dpm+ & 59.6 & 56.6 & 29.8 & 20 & 31.1 & 55.8 & 42.8 & 43.3 & 18.3 & 35.6 & 28.5 & 39.7 & 56.3 & 59.7 & 51.1& 19.6 & 42.1 & 33.1 & 49.1 & 50.3 & 41.1 \\
        conv-dpm+FT-all & 63.3 & 60.2 & 33.4 & 24.4 & 33.6 & 60 & 44.7 & 49.3 & 19.4 & 36.6 & 30.2 & 40.7 & 57.7 & 61.4 & 52.3 & 21.2 & 44.4 & 37.9 & 51.1 & 52.2 & {\bf 43.7} \\
        \hline
\end{tabular}
} 
\end{center}
\caption{Mean AP on PASCAL VOC 2011}
\label{tab:pascal11-results}
\end{table*}

\newpage
{
\bibliographystyle{ieee}
\bibliography{egbib}
}
\end{document}